\documentclass{amsart}
\usepackage{amsmath}%
\usepackage{amsfonts}%
\usepackage{amssymb}%
\usepackage{graphicx}

\usepackage{multirow}%
\usepackage{amsmath,amssymb,amsfonts}%
\usepackage{amsthm}%
\usepackage{mathrsfs}%
\usepackage[title]{appendix}%
\usepackage{xcolor}%
\usepackage{textcomp}%
\usepackage{manyfoot}%
\usepackage{booktabs}%
\usepackage{algorithm}%
\usepackage{algorithmicx}%
\usepackage{algpseudocode}%
\usepackage{listings}%

\theoremstyle{plain}

\numberwithin{equation}{section}

\begin{document}

\title[NN Learning of Black-Scholes Equation for Option Pricing]{Neural Network Learning of Black-Scholes Equation for Option Pricing}
\author{Daniel de Souza Santos}
\address[Daniel de Souza Santos]
{Education Department, Federal Institute of Education, Science and Tecnology.\newline%
\indent Rodovia PE 320, KM 126, Zona Rural. Serra Talhada, Caixa Postal 78, CEP 56915-899, Pernambuco, Brazil.}%
\email[Daniel de Souza Santos]{daniel.souza@ifsertao-pe.edu.br}%
\author{Tiago Alessandro Esp\' inola Ferreira}
\address[Tiago Alessandro Espínola Ferreira]
{Department of Statistical and Informatics, Federal Rural University of Pernambuco.  \newline%
\indent }%
\curraddr[Tiago Alessandro Espínola Ferreira]{Department of Statistical and Informatics, Federal Rural University of Pernambuco, Rua Dom Manoel de Medeiros, s/n, Dois Irm\~ aos, Recife, CEP 52171-970, Pernambuco, Brazil.\newline%
\indent }%
\email[Tiago Alessandro Espínola Ferreira]{tiago.espinola@ufrpe.br}%

\date{May 08, 2024}

\begin{abstract}
One of the most discussed problems in the financial world is stock option pricing. The Black-Scholes Equation is a Parabolic Partial Differential Equation which provides an option pricing model. The present work proposes an approach based on Neural Networks to solve the Black-Scholes Equations. Real-world data from the stock options market were used as the initial boundary to solve the Black-Scholes Equation. In particular, times series of call options prices of Brazilian companies Petrobras and Vale were employed. The results indicate that the network can learn to solve the Black-Sholes Equation for a specific real-world stock options time series.  The experimental results showed that the Neural network option pricing based on the Black-Sholes Equation solution can reach an option pricing forecasting more accurate than the traditional Black-Sholes analytical solutions. The experimental results making it possible to use this methodology to make short-term call option price forecasts in options markets.
\end{abstract}
\maketitle

\section{Introduction}

Differential equations modeling is employed for various scientific and engineering problems, describing the relationships between variables and their rates of change. Traditionally, solving these equations required complex analytical or numerical techniques. For many real-world problems, these differential equations are analytically intractable.  However, Artificial Neural Networks (or simply Neural network -- NN) have opened up exciting new possibilities for solving differential equations efficiently and accurately \cite{Lagaris1998, Khakifirooz2023, Zhu2023}.

Nowadays, it is possible to find some works in the literature where an NN is employed for a differential equations modeling problem. For instance, Uddin \cite{Uddin2023} uses wavelets as an activation function in a PINN (Physics-Informed Neural Networks) to solve the Blasius viscous flow problem. In this same problem, coupled linear differential equations, non-linear differential equations, and partial differential equations are solved. The problems solved are considered simple, but the approach has shown promise for solving more complex propositions. In another recent work, presented by Fang \cite{Fang2023}, a neural network was used to solve modified diffusion equations. The neural network was based on a mixture of Cartesian grid sampling and Latin hypercube sampling. They observed high accuracy when they compared the neural network results with other numerical solutions. They generalized the solver developed in their work to other partial differential equations. 

In Yang's \cite{YANG2023279} work, a network of stochastic differential equations induced by L\' evy was proposed to model complex time series data and solve the problem through neural networks. The methodology was applied to financial time series (agricultural products and equity indices). The researchers' team realized that the accuracy of predictions increased when non-Gaussian L\' evy processes were used. They also demonstrated that the proposed method's numerical solution converges in probability to the solution of the corresponding stochastic differential equation.

Another work was that of \cite{NOORANI2022112769} which suggests a new method to estimate the uncertain parameters of the inventory model led by the Liu process. First, an optimized artificial neural network was implemented based on actual data. Then, the estimation of the model's uncertainty parameters according to the optimized artificial neural networks was carried out. Nelder-Mead algorithm was used for the optimization of the ANN and the problem of estimating parameters. The main supremacy of the presented method was to provide a comparative algorithm and to demonstrate that the proposed approach can be effective for nonlinear problems. In this way, many other works \cite{Siegel2021, Khakifirooz2023, Zhu2023, Liu2023, Shin2023} also corroborate the idea of the NN usability to solve differential equations. 

When we focus on economics science applications, the differential equations appear in many finance problems \cite{NJIKETCHAPTCHET2023241, NAJAFI2022112875, KRAFT2023451}. One of these problems in finance is option pricing. In the 1970s, the Black-Sholes model\cite{black1973pricing} was developed. It proposed an analytical solution of a differential equation for a European option fair value calculation. This model was described by a second-order Parabolic Partial Differential Equation. Through some mathematical transformations, it is possible to show that the Black-Scholes differential equation can be rewritten as the heat equation, highlighting that the option pricing dynamic is similar to heat dissipation. 

The present work uses an MLP neural network to solve the Black-Scholes equations. With this NN, it is possible to create a forecaster for option pricing, where the NN is trained to solve the Black-Scholes model constrained real-world options data. The experiments on the stocks of two Brazilian blue chip companies (Petrobras and Valve) are used to demonstrate the NN capability to solve the Black-Scholes equation in a real-world situation. The NN's results are compared with real market data, where it is possible to observe the NN's ability to solve a real financial situation modeled by Black-Scholes equations.

This article is organized as follows. Section \ref{sec:Theoretical} shows the definitions and theoretical background for option pricing. Section \ref{sec:Methodology} presents the methodology proposed to solve the Balck-Sholes equation with an NN. Section \ref{sec:experimental}, the experimental setup is described, and in Section \ref{sec:results} the results are shown and discussed. Finally, Section \ref{sec:Conclusions} exhibits the conclusions about the work.    

\section{Theoretical Definitions}\label{sec:Theoretical}

\subsection{What is an Option?}

In the financial market, contracts have been developed to trade assets (stocks or commodities, for example) for a future date and at a price set in the contract. These contracts are called derivatives \cite{AMBROSELOBOOK}. 

The present work only covers derivatives traded on the stock exchange. In particular, on the Brazilian market B3 (in Portuguese: \textit{\underline{B}rasil, \underline{B}olsa, \underline{B}alc\~ ao}).

According to Hull~\cite{GVK563580607}, derivatives have been developed as a financial instrument for transferring risks not intrinsic to the economic activity that the producer is engaged to another party interested in taking such risk by receiving a remuneration.

The main types of derivatives are Forwards, Futures, Options, and Swaps. One of the most widely used derivatives in the stock market is options. There are two types of options, named call options and put options. A call option gives its holder the right, not the obligation, to buy a particular asset at a given price at a certain future date. The put option buyer must sell the asset at the correct price. A put option gives its holder the right to sell an asset at a certain price on a specific date. The writer of the put option, \textit{i.e.} the person who made the put available on the market and sold it, is obliged to purchase the underlying asset at the price agreed on the combined date \cite{AMBROSELOBOOK}.

There are still two other classifications for the options: American and European. American options can be exercised at any time until the expiration of the option. European options can only be exercised at a certain date \cite{GVK563580607}. In the Brazilian market, the options are usually exercised on the third Friday of each month.

There are several recent studies on pricing options, such as the work of \cite{Cao2023, Monteiro2023, Nabubie2023, Oh2023}. The pricing calculation of an American option can only be done numerically, and several studies are looking for the best model for calculating the fair price of an American option \cite{Shirzadi2023, Lee2021, Mehrdoust2021, Yan2022, Gyulov2022, Zaevski2022}. However, here we are interested only in the European options.

Let us take an example: consider the stocks of an ABC company. At the time this text is being written, the value of a stock is 35.57 USD. An investor can write a 30.00 USD Strike European call option with expiration within 90 days. If another investor decides to buy this option, he or she will have the right to exercise it or not on the scheduled date. If ABC stocks are traded at 40.00 USD in 90 days, the call owner can exercise it and buy the stocks worth 40.00 USD for 30.00 USD and sell them immediately afterward, earning a profit of 10.00 USD per share. The writer of the call option will need to buy shares for 40 USD on the market and sell them immediately for 30 USD to the holder of the options he has issued, if he does not own these shares in his investment portfolio. On the other hand, if ABC stocks are being traded at 20.00 USD in 90 days, it makes no sense for the call holder to buy an equity that is worth 20.00 USD for 30.00 USD, then it is said that the option has turned dust, it is worth approximately 0.00 USD. The call writer will get the amount paid by the buyer of the option and s/he will be a profit. On the other hand, the buyer of the option will have lost the money invested in the call option on ABC stocks.

The call payoff is the difference between the strike price of $K$ and the current price of the stock $S(t)$, commonly referred to as the spot price. If the result is positive, the holder of the option can exercise it, buy stocks for the value of $K$, and sell it for $S(t)$ making a profit. If the result is negative, the option value is zero and the call buyer loses the amount invested. The payoff for a long position on a European call option \footnote{long position is the same as the option holder, short position is the same as option writer or seller of options.} at the expiration time $T$, $S(t = T)$, is given by,

\begin{equation}
	\mathop{\mathrm{payoff}} = \max[S(T) - K, 0].
	\label{payoffEquation}
\end{equation}

As an example of a put option, consider the BCD company stocks. At the time this text is being written, BCD stocks are being traded at 38.51 USD. Imagine an investor wishing to issue a European put option on these stocks. Consider as a first case that the put option has a strike of 45.00 USD and expires within 180 days. If in 180 days the BCD stocks are traded at 25.00 USD, the put writer will have to sell it for 45.00 USD, even if it is being traded at 25.00 USD, that is, it will take a loss. However, if BCD is being traded at 50.00 USD, the option writer can stay calm and save his profit. The put owner will prefer to sell the stock to the market (50.00 USD) than for 45.00 USD, which was the right price in the option trade. 

It can be understood that put options function as a hedging tool for the owner's assets. For more information about hedge strategies using selling options, see the works \cite{Anderegg2022, Borochin2021, Brigatto2022, Chi2023, Cho2022, Chung2013, Daniliuk2015, Nian2021}.

In real-world data, the options are labeled following different rules. Looking at the Brazilian market, the following nomenclature rule is used for options: the first four letters refer to the name of the underlying asset. The fifth letter indicates whether it is a call option (from A to L) or a put option (from M to X). After that, there are two or three numbers, to form the complete Brazilian Market Option codification.  For the present study, 12 price series for PETRA, 11 price series for PETRD, and 10 price series for VALED were used.

\subsection{Black-Scholes}

The Black-Scholes European options pricing model is based on a second-order Parabolic Partial Differential Equation. For a stock that does not pay dividends, we have \cite{GVK563580607}:

\begin{equation}
	\frac{\partial c}{\partial t} + rS \frac{\partial c}{\partial S} + \frac{1}{2}\sigma^2S^2\frac{\partial^2c}{\partial S^2} = rc
	\label{BlackScholesEquation}	
\end{equation}

where $t$ is the time, $c$ is the call option price (or change it to $p$ for a put call), $S$ is the Spot stock price, $r$ is the free interest rate, and $\sigma$ is the stocks' volatility. In the Brazilian market, $r$ is named SELIC. For all experiments done,  the real SELIC data always was a constant, with a value of $13,75\%$ per year. 

The solutions to Equation \ref{BlackScholesEquation} are the Black-Scholes-Merton formulas for European call and put options pricing. The formulas are \cite{BlackFischer1973TPoO}: 

\begin{equation}\label{eq:CallSolution}
	c = S_0 N(d_1) - K\exp(-rT)N(d_2)
\end{equation}
and:
\begin{equation}
	p = K\exp(-rT)N(-d_2)-S_0N(-d_1)
\end{equation}

where:

\begin{equation}
	\begin{cases}
		d_1 = \dfrac{\ln\left(\dfrac{S_0}{K}\right) + \left(r+\dfrac{\sigma^2}{2}\right)(T-t)}{\sigma \sqrt {T - t}}
		\\
		d_2 = \dfrac{\ln\left(\dfrac{S_0}{K}\right) + \left(r-\dfrac{\sigma^2}{2}\right)(T - t)}{\sigma \sqrt {T - t}} = d_1 - \sigma \sqrt {T - t}
	\end{cases}\,.
\end{equation}

The $N(x)$ function is the Cumulative Distribution Probability function for a random variable with a standardized normal distribution. The variables $c$ and $p$ are the call and put prices, respectively. $S_0$ is the stock price at zero time, $K$ is the strike price and $T$ is the time until the option maturity.

Analytical solutions are only possible for European options, that can only be exercised on a specific date. There is still no analytical model for calculating the fair price of American options. It is possible to solve the Black-Scholes equation analytically in various ways, such as through a binomial tree \cite{GVK563580607}, using Hermites' polynomials \cite{XIU2014158}, or transforming it into another differential equation \cite{ROMAN2004}. Here, the adopted approach was to transform the Black-Sholes equations into an alternative differential equation. 

Two boundary conditions have been used for this problem. The first boundary condition was the payoff, explained earlier. Rewriting the equation \ref{payoffEquation}, the mathematical equation that represents the payoff of a call is \cite{RITELLIBOOK}:
\begin{equation}
	u(x, 0) = \max\left\{\exp\left[\dfrac{1}{2}(k+1)x\right] - \exp\left[\dfrac{1}{2}(k-1)x\right],0\right\}.
\end{equation}
where $k = \dfrac{2r}{\sigma^2}$.

This boundary condition is the data modified by the following equations:

\begin{equation}
	x = \ln \left[\dfrac{S(t)}{c(t)} \right]
    \label{eq:x}
\end{equation}
and
\begin{equation}
	\tau = \frac{1}{2}\sigma^2(T - t)
    \label{eq:tau}
\end{equation}
where $T$ is the strike time.

To change variables, the function $c(s, t)$ must first be mapped to the function $f(x, \tau)$, using the Strike $K$ price as a scale factor:

\begin{equation}
    c(s,t) = Kf(x, \tau).
    \label{eq:FirstChangeVariable}
\end{equation}

Then we should do another mapping in a function $u$, but this time, keeping the variables $x$ and $\tau$:

\begin{equation}
    f(x, \tau) = \exp\left({\alpha x + \beta \tau} \right) u(x,\tau).
    \label{eq:SecondChangeVariable}
\end{equation}

After calculating the values of $\alpha$ and $\beta$, and using the equations \ref{eq:FirstChangeVariable} and \ref{eq:SecondChangeVariable}, we get a direct relationship between the actual price of the option $c(s, t)$ and the modified price $u(x, \tau)$, which will be used in the resolution of the heat equation by the ANN:

\begin{equation}
    c (s, t) = K u(x, \tau) \exp{\left[ -\dfrac{1}{2}(k-1)x - \dfrac{1}{4}(k+1)^2 \tau \right]}.
    \label{eq:c2u}
\end{equation}

It is possible to demonstrate that the Black-Scholes equation can be rewritten as the heat transmission equation on a metal bar \cite{RITELLIBOOK}. 

\begin{equation}
	\frac{\partial u}{\partial \tau} = \frac{\partial^2 u}{\partial x^2},\; -\infty < x < \infty, \;\tau > 0.
	\label{HeatEquation}
\end{equation}

Therefore, the Equation \ref{BlackScholesEquation} became the Equation \ref{HeatEquation}, where $u$ is the generalized price, \textit{i.e.} the price of the option after the mathematical transformations. The relationship between the generalized price $u$ and the real-world price $c$ is given by the Equation \ref{eq:c2u}. The $\tau$ is the generalized unit of time, calculated using Equation \ref{eq:tau} and the $x$ is the generalized price of the share, derived from Equation \ref{eq:x}.

Options price data can be obtained free of charge on the ADVFN website\footnote{\textit{http://https://br.advfn.com/}}. Options were selected concerning the two most traded stocks on the Brazilian Stock Exchange: Petrobras (PETR4) and Vale (VALE3). There are options from other traded companies, such as Banco do Brasil (BBAS3) and WEG (WEGE3). However, their trading volume is very low, which could compromise data analysis because there are many days without trading operations for these two companies. 

As a case study, given data availability limitations, only the call option series were used for the NN training. The puts did not have significant trading volume.

\section{Methodology}\label{sec:Methodology}

To solve a differential equation using a neural network, we can treat it as an optimization problem. Let $\mathcal D (\cdot)$ be a differential operator and $u$ a possible solution of $\mathcal D (\cdot)$. Consider a differential equation in the form:

\begin{equation}
	\mathcal D (u) - \mathcal F = 0.
\end{equation}
where $\mathcal F$ is a known forcing function.

Let $\hat u$ the NN output and  whether $\hat u$ is a trial solution for the differential equation $\mathcal D (\cdot)$, then the residual $\mathcal R (\hat u)$ is:

\begin{equation}\label{eq:residuo}
	\mathcal R (\hat u) = \mathcal D (\hat u) - \mathcal F.
\end{equation}

In this way, an NN can be trained to optimize the solution $\hat{u}$ with a loss function given by Equation \ref{eq:residuo}. The solving differential equation problem is reduced to a minimization problem.

To guarantee that the initial conditions are satisfied, the function $\hat u$ can be changed to the modified solution $\Tilde{u}$. For example, if a given differential equation in space $x$ and time $t$ has a initial condition in $t=t_0$ given by the function $u_{t_0}(x)$, the solution can be written as

\begin{equation}
    \tilde{u}(x,t) = u_{t0}(x)+\left[ 1-e^{-(t-t_0)} \right] \hat{u}(x,t)
\end{equation}

In general form, many other initial conditions can be implemented in the form, 
\begin{equation}
    \tilde{u}(x,t) = A(x,t;x_{boundary},t_0)\hat{u}(x,t)
\end{equation}
where $A(x,t;x_{boundary},t_0)$ is selected so that $\tilde{u}(x,t)$ has the correct initial and boundary conditions. All these conditions implementation are found in the Neurodiffeq library \cite{Chen2020}.

With the NN procedure to solve differential equations, the idea is to apply NN to solve the Black-Scholes Equation.  The previous section shows the Black-Scholes Equation (\ref{BlackScholesEquation}) and its equivalent, the heat version equation (\ref{HeatEquation}). The data used for the network training were the price series of two options of the Brazilian market: Petrobras and Vale. These price series correspond to maturity in January (series A: PETRA) and in April (Series D: PETRD and VALED). The same mathematical transformations applied to the Black-Scholes Equation were applied to the price data of options. The details of these mathematical transformations will be presented in Section \ref{sec:experimental}.

\section{Experimental setup}\label{sec:experimental}

The Neurodiffeq Python library ~\cite{Chen2020} was employed to train the ANN to solve the Black-Scholes' heat version equation (Equation \ref{HeatEquation}). All computational simulations were implemented in Python 3 programming language and used the Torch framework for the neural network. The neural network parameters were: 
\begin{itemize}
    \item Activate Function: Hyperbolic Tangent;
    \item Training algorithm: Adam
    \item NN Architecture: an MLP with 2 inputs, two hidden layers with 32 neurons, and an output (2-32-32-1);
\end{itemize}

In particular, the NN architecture was adapted to the problem, since the equation involves two variables, one represents a modified price ($x$) and the other represents an interval of time ($\tau$). Thus, the network has two inputs. The network output is just a neuron, which represents the option price (the Black-Scholes solution).

The mathematical transformations were applied to the price data of the options (PETRA, PETRD, and VALED). After that, these were used as boundary conditions. The NN was trained 30,000 times (epochs), and the error measurements included MAE, MSE, MAPE, POCID, and ARV. These metrics are described in the next section. 

\subsection{Error metrics}

\subsubsection{Mean Absolute Error (MAE)}

It is a measure of absolute deviations between the actual and predicted points. It is calculated as the absolute value of the difference between the actual values $(Y)$ and the estimated values $(\hat Y)$, over the sample size $(N)$. As with other measurements of error, the closer to zero, the lower the error of the estimate and the better the model performance.

\begin{equation}
	MAE = \dfrac{1}{N} \sum\limits_{i = 1}^{N} |Y_i - \hat Y_i|
\end{equation}

\subsubsection{Mean Square Error}

It is the mean square deviation between the actual and predicted points. It is the square difference between the actual values $(Y)$ and the estimated values $(\hat Y)$, normalized by the sample size $(N)$. Since the differences are square, it always results in positive values. The closer the MSE is to zero, the lower the error associated with the measurements.

\begin{equation}
	MSE = \dfrac{1}{N}\sum_{i = 1}^{N} (Y_i - \hat Y_i)^2
\end{equation}

\subsubsection{Mean Absolute Percentage Error (MAPE)}

It is also a measure of precision. It is the difference between estimated values $(\hat Y)$ and actual values $(Y)$, divided by the actual value, in the module. The sum of these proportions is divided by the size of the sample $(N)$. The closer to zero, the less the error of the estimate.

\begin{equation}
	MAPE = \dfrac{1}{N} \sum\limits_{i = 1}^{N} \left| \frac{Y_i - \hat Y_i}{Y_i}\right|
\end{equation}

\subsubsection{Prediction of Change in Direction (POCID)}

When forecasting whether the value of the series will increase or decrease in the upcoming time steps, the Prediction of Change in Direction (POCID) measure enables an accounting of the number of accurate decisions. Mathematically \cite{ARTICLETAEF}:

\begin{equation}
POCID = 100\dfrac{\sum\limits_{i = 1}^ND_i}{N}
\label{eq:POCID}
\end{equation}

where
\begin{equation}
D_i = \begin{cases} 1, & \mbox{if }  (Y_i - Y_{i-1}) \times (\hat{Y}_i - \hat{Y}_{i-1})\\ 0, & \mbox{Otherwise.}  \end{cases}
\label{eq:D}
\end{equation}

The POCID value can vary between 0 to 100\%, where the perfect model reaches the value of 100\%.

\subsubsection{Average Relative Variance (ARV)}

The last relevant evaluation measure is the Average Relative Variance (ARV) \cite{ARTICLETAEF}:

\begin{equation}
	ARV = \dfrac{1}{N} \dfrac{\sum\limits_{i = 1}^N(\hat{Y}_i - Y_i)^2}{\sum\limits_{i = 1}^N(\hat Y_i - \overline{Y})^2}
\label{eq:ARV}
\end{equation}
where $N$, $Y$, and $\hat Y$ are the same parameters of the other evaluation measures, and $\overline{Y}$ is the time series mean. When the ARV value is one, the predictor performs the same as if it were the mean of the series; when the value is more than one, it performs worse than if it were the mean; and when the value is less than one, it performs better than if it were the mean. Hence, the predictor tends to be the ideal model when the ARV decreases to zero and is useful if the value of the ARV is less than 1.

\section{Results}\label{sec:results}

We use an MLP Neural Network to solve a supervised learning problem: the resolution of a second-order Parabolic Partial Differential Equation, namely the Black-Scholes' heat version equation. For their training, we used data from the Brazilian market options, the options on Petrobras and Vale companies. The options were from two series classes, named A and D. The A series maturity occurred in January 2023, and the D series maturity occurred in April 2023. The results of the NN forecasting error metrics can be found in Tables \ref{tab:PETRA}, ~\ref{tab:PETRD}, and \ref{tab:VALED}. The tables also show the $N$ size of the price series. 

\begin{figure}[!h]
		\begin{center}
			\includegraphics[width=0.48\linewidth]{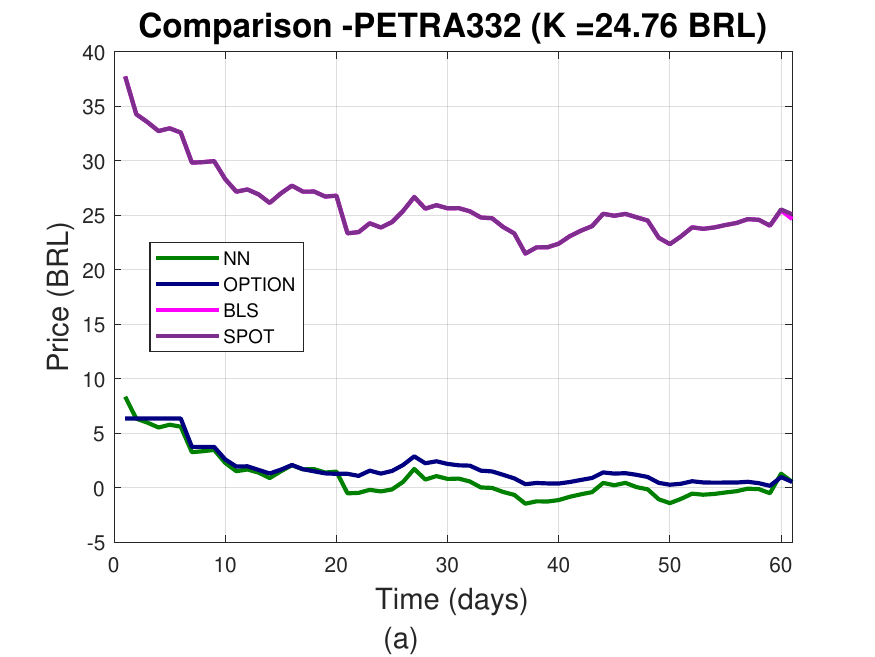} \quad
            \includegraphics[width=0.48\linewidth]{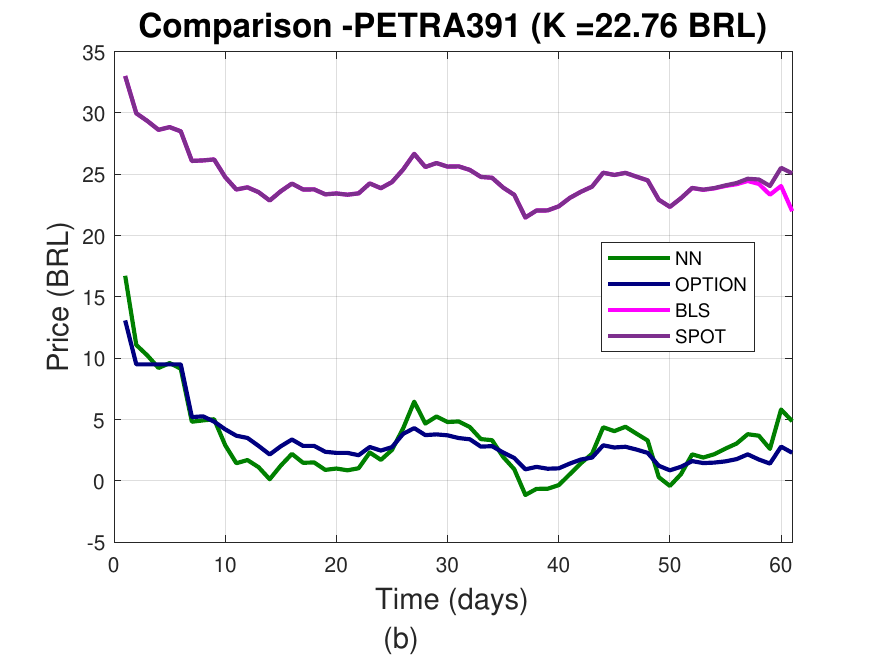} \\
			\includegraphics[width=0.48\linewidth]{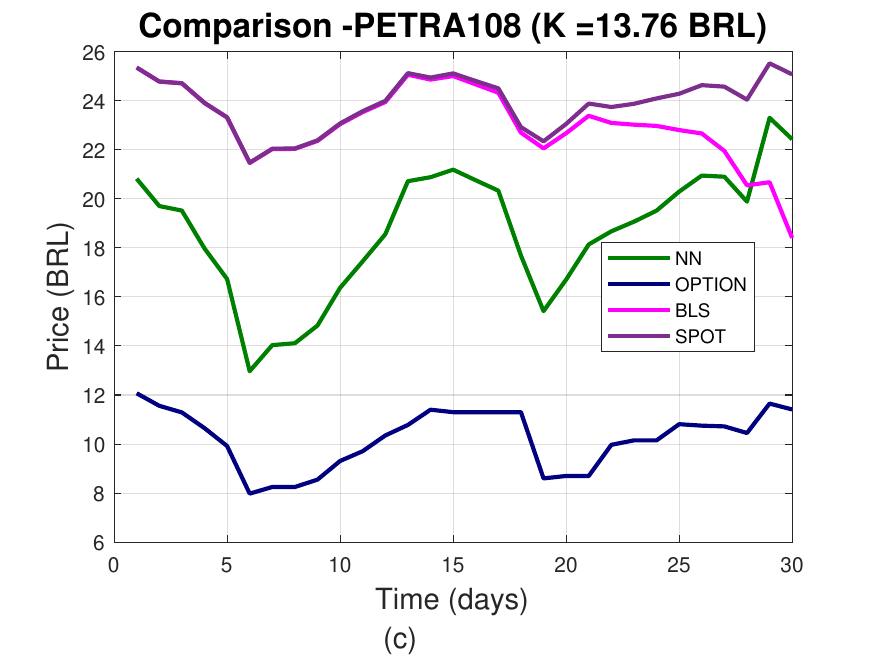} 
   
			\caption{(a) Results for PETRA332 (MSE minor) (b) Results for PETRA391 and (c) results for PETRA108 (MSE major). (a) and (b) are the best NN prediction cases, and (c) is the worst NN prediction case. For all graphics, there are four curves: the price of the underlying PETR4 stock - OPTION, the blue curve; the price calculated by the  Black-Scholes analytical solution - BLS, the magenta curve; the price computed by the ANN, the green curve; and, the option market price - SPOT, the purple curve.} 
			\label{fig:PETRA}
		\end{center}
	\end{figure}

Table \ref{tab:PETRA} provides error statistics for the PETRA call options. The best results for each statistics error measure are highlighted in boldface. The NN prediction for the call option PETRA332 ($K = 24.76$ BRL) showed the lowest values of MAE and MSE (0.980 and 1.295, respectively). The PETRA391 option ($K = 22.76$ BRL) showed the lowest MAPE value ($0.528$) and the lowest ARV ($0.003$). Of the five analytical metrics used, these two were better in two of them. Even PETRA391 showed a POCID value slightly higher than PETRA332 ($78.333$ of this versus $75.000$ of that), looking at the charts (Figures \ref{fig:PETRA}a and \ref{fig:PETRA}b), it was possible to see that the PETRO332 option had its values closest to the values recorded in the market. We also can see in Figure ~\ref{fig:PETRA} the behaviors of the series with lower and higher MSE concerning the real price series and the solution of the Black-Scholes equation for the data presented. Four curves are presented, the price of the underlying stock, which in Figure ~\ref{fig:PETRA} case is PETR4 (SPOT - purple line), the price calculated by the analytical solution (Equation~\ref{eq:CallSolution}) of the Black-Scholes Equation (BLS - magenta line), option market price (OPTION - blue line) and the price computed by the NN (green line). The closer the green line is to the blue line, the better the results. For comparison purposes, we also chose the largest MSE to demonstrate the worst price behavior, and in the case of series A, it was PETRA108 ($K = 13.76$, $MSE = 73.457$), whose graph is also shown in Figure \ref{fig:PETRA}.

One general phenomenon observed here was the behavior of Black-Scholes' analytical solution. For dates far from maturity, it converges in the share price (SPOT). When the time comes to option maturity, the analytical solution tends to get closer to the actual option value.

\begin{table}[!h]
	\caption{Statistical Errors for MLP Neural Network Modeling the Petrobras Options with Black-Scholes Model - A Series for PETR4. The best value for each error measure is in boldface.}
	\label{tab:PETRA}
\begin{tabular}{c|ccccccc}
\hline
\hline
\textbf{PETRA} & \textbf{K (BRL)} &\textbf{MAE} & \textbf{MSE} & \textbf{MAPE} & \textbf{POCID} & \textbf{ARV} & \textbf{N}\\ 
\hline
\textbf{332}   & 24.76  & \textbf{0.980}  & \textbf{1.295}  & 1.257  & 75.000  & 0.004 & 61  \\
\textbf{357}   & 26.26  & 1.145  & 1.641  & 2.590  & 77.358  & 0.023 & 54  \\
\textbf{356}   & 25.76  & 1.345  & 2.064  & 2.733  & 82.353  & 0.020 & 52  \\
\textbf{366}   & 26.76  & 1.141  & 2.337  & 2.114  & 71.667  & 0.020 & 61  \\
\textbf{342}   & 25.26  & 1.530  & 2.706  & 2.357  & \textbf{84.314}  & 0.019 & 52  \\
\textbf{266}   & 17.76  & 5.749  & 37.778 & 0.800  & 53.333  & 0.013 & 61  \\
\textbf{391}   & 22.76  & 1.179  & 1.888  & \textbf{0.528}  & 78.333  & \textbf{0.003} & 61  \\
\textbf{321}   & 22.26  & 1.670  & 4.563  & 0.533  & 71.667  & 0.006 & 61  \\
\textbf{282}   & 18.76  & 6.231  & 42.81  & 1.084  & 72.340  & 0.019 & 48  \\
\textbf{53}    & 15.26  & 2.800  & 10.538 & 0.598  & 46.154  & 0.023 & 40  \\
\textbf{108}   & 13.76  & 8.418  & 73.457 & 0.820  & 72.414  & 0.032 & 30  \\
\textbf{362}   & 27.26   & 1.011 & 1.942  & 2.163  & 68.333  & 0.020 & 61  \\
\hline
\hline                           
\end{tabular}
\end{table}

\begin{figure}[!h]
		\begin{center}
			\includegraphics[width=0.48\linewidth]{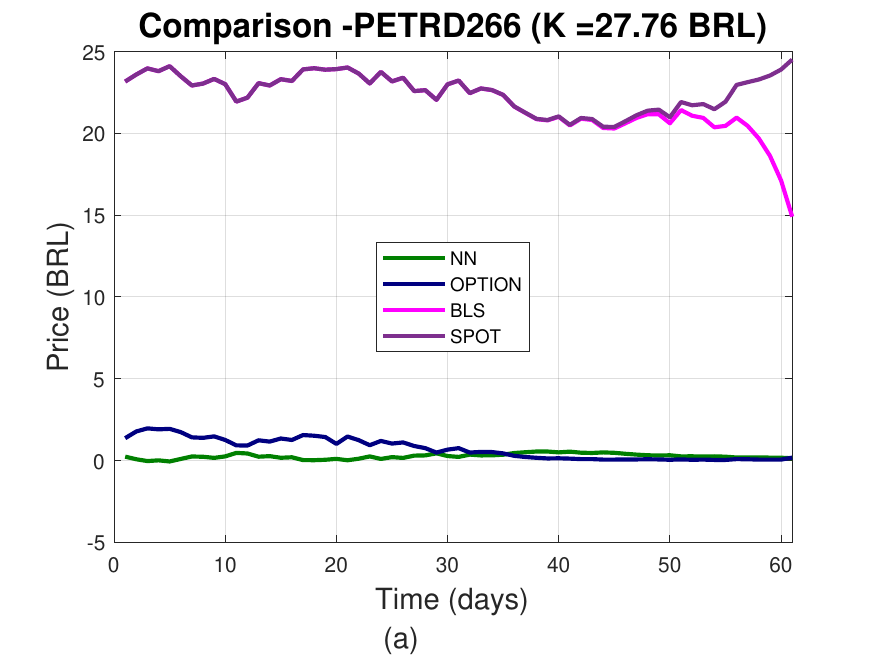} \quad
			\includegraphics[width=0.48\linewidth]{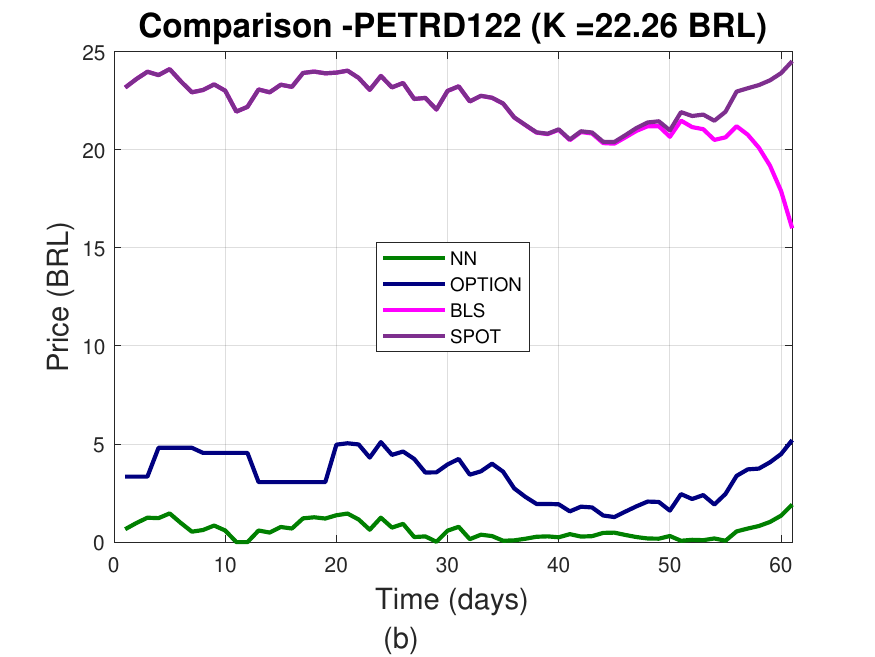} \\
            \includegraphics[width=0.48\linewidth]{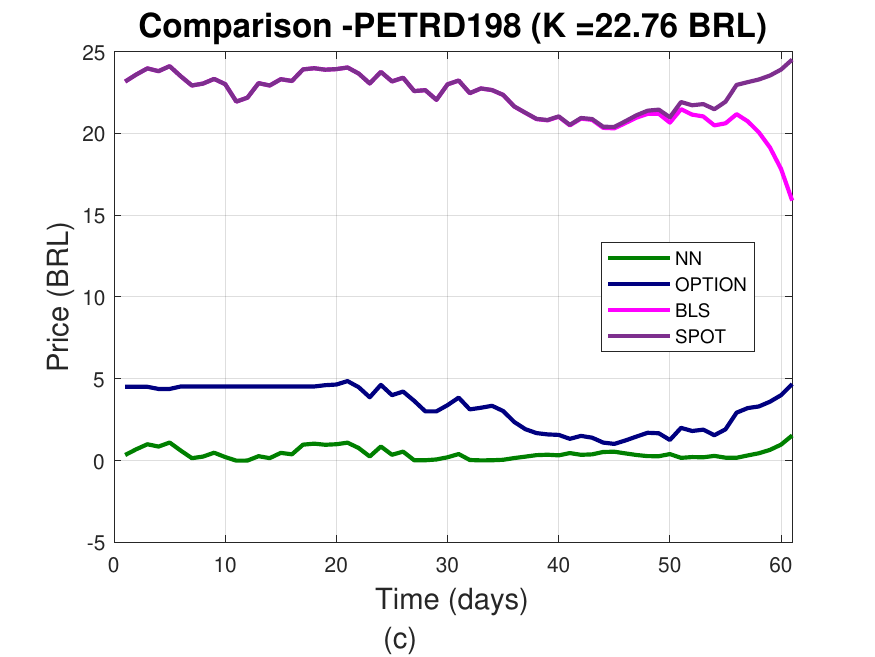}
			\caption{(a) Results for PETRD266 (MSE minor), (b) Results for PETRD122 and (c) results for PETRA198 (MSE major). (a) and (b) are the best NN prediction cases, and (c) is the worst NN prediction case. For all graphics, there are four curves: the price of the underlying PETR4 stock - OPTION, the blue curve; the price calculated by the  Black-Scholes analytical solution - BLS, the magenta curve; the price computed by the ANN, the green curve; and, the option market price - SPOT, the purple curve.} 
			\label{fig:PETRD}
		\end{center}
	\end{figure}

Table \ref{tab:PETRD} shows the error metrics results for the PETRD set of options, \textit{i.e.} options expiring in April 2023. One more time, the best results for each statistics error measure are highlighted in boldface. The NN price prediction for PETRD266 option ($K = 27.76$ BRL) showed the lowest values for MAE ($0.670$) and MSE ($0.751$) and the NN solution for call PETRD122 ($K = 22.26$ BRL) had the lower recorded value of ARV ($0.018$) and the highest POCID ($35.000$). Since both had the best values on two of the five indicators chosen, the choice for the best case was made based on the chart of the two options (Figures \ref{fig:PETRD}a and \ref{fig:PETRD}b). In this data set all series have the same size of $61$ points. POCID showed low values for all Petrobras D Series options. As in the case of PETRA, the decision on the chart to be chosen was made based on the MSE. The highest MSE occurred in PETRD198 ($9.541$), Figure \ref{fig:PETRD}c, whose strike price was $k=22.76$ BRL. 

Figure \ref{fig:PETRD} follows the same color scheme as Figure \ref{fig:PETRA}, and a similar Black-Scholes' analytical solution behavior, where the solution values are almost even the price of PETR4. However, one more time, near maturity, the Black-Scholes analytical solution (BLS) departs from the asset price and tends to approach the actual option price. The NN's numerical solution (green line) from the beginning of the option's life is closest to the actual option price (blue line). Even in the case of the largest MSE, PETRD198, the NN solution is much closer to the real value than BLS.

\begin{table}[!h]
	\centering
	\caption{Statistical Errors for MLP Artificial Neural Network (ANN) Modeling the Petrobras Options with Black-Scholes Model - D Serie for PETR4.  The best value for each error measure is in boldface.}
	\label{tab:PETRD}
\begin{tabular}{c|ccccccc}
		\hline
		\hline
\textbf{PETRD} & \textbf{K (BRL)} & \textbf{MAE} & \textbf{MSE} & \textbf{MAPE} & \textbf{POCID} & \textbf{ARV} & \textbf{N} \\
		\hline
\textbf{291}    & 28.26 & 0.727        & 0.906        & 1.922         & 16.667         & 0.028        & 61  \\
\textbf{222}    & 26.76 & 1.114        & 1.664        & 3.324         & 8.333          & 0.069        & 61  \\
\textbf{266}    & 27.76 & \textbf{0.670}& \textbf{0.751}        & 2.115         & 11.667         & 0.054        & 61  \\
\textbf{271}    & 27.26 & 0.806        & 1.020        & 2.507         & 13.333         & 0.067        & 61  \\
\textbf{220}    & 26.26 & 1.198        & 1.997        & 2.231         & 10.000         & 0.056        & 61  \\
\textbf{227}    & 25.26 & 1.539        & 3.602        & 0.938         & 10.000         & 0.030        & 61  \\
\textbf{202}    & 25.76 & 1.301        & 2.484        & 1.377         & 6.667          & 0.040        & 61  \\
\textbf{194}    & 23.76 & 2.173        & 6.636        & \textbf{0.771}         & 16.667         & 0.024        & 61  \\
\textbf{183}    & 24.76 & 1.798        & 4.599        & 1.089         & 11.667         & 0.036        & 61  \\
\textbf{198}    & 22.76 & 2.848        & 9.541        & 0.855         & 33.333         & 0.019        & 61  \\
\textbf{122}    & 22.26 & 2.751        & 8.501        & 0.822         & \textbf{35.000}         & \textbf{0.018}        & 61  \\ 
		\hline
		\hline     
\end{tabular}
\end{table}

\begin{figure}[tb]
		\begin{center}
			\includegraphics[width=0.48\linewidth]{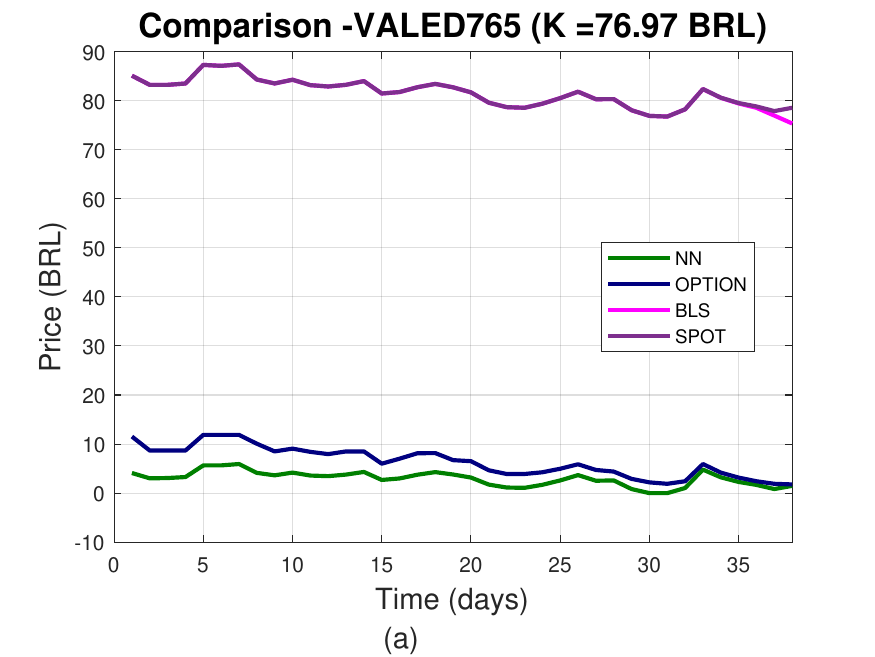} \quad
			\includegraphics[width=0.48\linewidth]{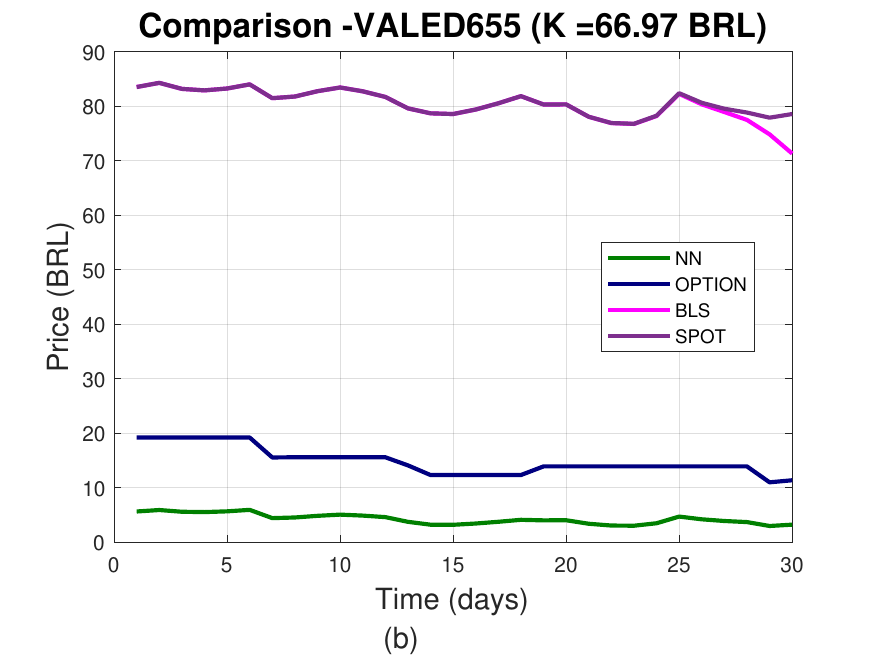}
			\caption{(a) Results for VALED765 (MSE minor, the bets NN prediction) and (b) results for VALED655 (MSE major, the worse NN prediction). For both graphics, there are four curves: the price of the underlying PETR4 stock - OPTION, the blue curve; the price calculated by the  Black-Scholes analytical solution - BLS, the magenta curve; the price computed by the ANN, the green curve; and, the option market price - SPOT, the purple curve.} 
			\label{fig:VALED}
		\end{center}
	\end{figure}

Table \ref{tab:VALED} provides error statistics for VALE3 call options. Again, the best results for each statistics error measure are highlighted in boldface. The NN option prediction with lower values of MAE and MSE ($3.435$ and $15.172$, respectively) occurred to VALED765 series ($K = 76.97$ BRL). This NN option predition also showed the highest value of POCID, $78.378$, which means that NN accords $78\%$ of price movements. The NN predition with lowest ARV  ($0.019$) occurred for VALED80 series ($K = 78.97$ BRL). The NN prediction with lowest MAPE ($0.484$) occurred for VALED75 series ($K=73.97$ BRI). Figure \ref{fig:VALED}a shows the behavior of the VALED765 price series, which had the lowest value of MSE and MAE. Figure \ref{fig:VALED}b shows results for VALED655 series ($K = 66.97$ BRL), which showed the highest value for MSE ($115.290$). 

Observing Figure \ref{fig:VALED}, it is clear that even in the case of higher MSE, the NN numerical solution (green line) is much closer to the option real values (blue line) than the Black-Sholes analytical solution (magenta curve).

\begin{table}[!ht]
\centering
\caption{Statistical Errors for MLP Neural Network Modeling the Vale Options with Black-Scholes Model - D Series for VALE3. The best value for each error measure is in boldface.}
\label{tab:VALED}
\begin{tabular}{c|ccccccc}
		\hline
		\hline
\textbf{VALED} & \textbf{K (BRL)} & \textbf{MAE} & \textbf{MSE} & \textbf{MAPE} & \textbf{POCID} & \textbf{ARV} & \textbf{N} \\
\hline
\textbf{80}    & 78.97 & 5.164        & 32.565       & 0.792         & 58.333         & \textbf{0.019}        & 61 \\
\textbf{77}    & 75.97 & 4.539        & 26.032       & 0.547         & 77.778         & 0.025        & 46 \\
\textbf{765}   & 76.97 & \textbf{3.435}  & \textbf{15.172}       & 0.540         & \textbf{78.378}         & 0.029        & 38 \\
\textbf{79}    & 77.97 & 6.727        & 59.320       & 0.771         & 60.000         & 0.021        & 61 \\
\textbf{743}   & 74.97 & 4.213        & 22.747       & 0.510         & 71.429         & 0.033        & 36 \\
\textbf{81}    & 79.97 & 5.617        & 41.571       & 0.850         & 56.667         & 0.021        & 61 \\
\textbf{83}    & 81.97 & 4.901        & 35.771       & 0.951         & 73.214         & 0.025        & 57 \\
\textbf{82}    & 80.97 & 5.536        & 42.683       & 0.887         & 66.667         & 0.022        & 61 \\
\textbf{75}    & 73.97 & 4.814        & 29.249       & \textbf{0.484}         & 60.976         & 0.028        & 42 \\
\textbf{655}   & 66.97 & 10.604       & 115.290      & 0.715         & 20.690         & 0.034        & 30 \\ 
		\hline
		\hline      
\end{tabular}
\end{table}

		It can be observed that options that had a lower exercise price than the stock price on the expiration date (\emph{In The Money - ITM}) had higher trading volume and presented fewer estimate errors. 
		
A comparative performance analysis with other results in options is a very hard task. The objective and data are different when compared with the results presented here. However, a rouge performance baseline can be traced in Liang \textit{et. al.} work \cite{LIANG20093055}. They conducted an options price forecasting analysis for the Hong Kong market employing neural networks, Support Vector Regression, and other methodologies. The article displays the options for Hang Seng Bank and although the market is quite different from the Brazilian market, the time structure of the series is similar to the presented work here. The best scenario analyzed by Liang et al. presented an MAE of $14.3$ for forecasts with neural networks. Our work presented the best scenario when studying options on Petrobras, whose MAE was about seven times lower, around $2.6$. When looking at the VALE times series, our NN prediction reached an average MAE of $5.6$. Thus, our methodology presents better efficiency for the pricing of European options in a broad and general context.

\section{Conclusions}\label{sec:Conclusions}

One of the most discussed problems in the financial world is the calculation of the fair value of a stock option. This problem is the subject of several academic articles and there is still no consensus on which is the best method for price options. A special case is the European options, for which an analytical model was developed in the 1970s: the Black-Scholes model, which is the resolution of a Parabolic Particle Differential Equation of Second Order.  The consolidated knowledge of the fair price of a European option is precisely the analytical solution of the Black-Scholes equation, which has been demonstrated that this solution presents values very far from the real values practiced on the market for more distant maturity dates. However, based on the experimental results reached, the big problem is not in the Black-Sholes equation, but the real applications problems probably comes from the analytical solution employed. This article is focused on solving the Black-Scholes equation through an artificial neural network. This methodology is innovative because it starts from the same differential equation and manages to find a solution closer to the reality of the derivative market.

For the resolution of the equation, an MLP neural network was implemented with the Python 3 neurodiffeq library \cite{Chen2020}. This is a supervised learning problem in which the correct answer is given by actual data. Due to the low trading volume, the put options were not studied, the methodology being applied only to Petrobras and Vale call options in the Brazilian market. It can be seen that options with a lower strike price than the stock price have a higher trading volume. These options are called \textit{In The Money} (ITM). The neural network learned from the data, generated significantly low estimate errors, suggesting that this methodology is efficient for solving the Black-Scholes Equation. In the future, options from other companies in other options series will be evaluated. Estimates will also be made through arima modeling to evaluate the NN performance for purchase options price predictions.

\bibliographystyle{plain}
\bibliography{References}

\end{document}